%% file: main.tex
\documentclass[runningheads]{llncs}
\usepackage{graphicx}
\usepackage[toc,page,header]{appendix}
\usepackage{tikz}
\usepackage{comment}
\usepackage{amsmath,amssymb} 
\usepackage{color}
\usepackage{xcolor}
\usepackage{booktabs}
\usepackage{multirow}
\usepackage{subcaption}
\usepackage[accsupp]{axessibility}  
\usepackage{pifont}
\newcommand{\cmark}{\ding{51}}%
\newcommand{\xmark}{\ding{55}}%

\usepackage[numbers,sort,compress]{natbib}
\usepackage[pagebackref,breaklinks,colorlinks]{hyperref}

\begin{document}

\pagestyle{headings}
\mainmatter

\title{A CLIP-Hitchhiker's Guide to Long Video Retrieval}

\author{Max Bain$^{1}$ \and
Arsha Nagrani$^{1}$ \and
G\"ul Varol$^{2}$ \and
Andrew Zisserman$^{1}$}
\authorrunning{M. Bain et al.}
%
\institute{Visual Geometry Group, University of Oxford \and
LIGM, \'Ecole des Ponts, Univ Gustave Eiffel, CNRS
\email{\{maxbain,arsha,gul,az\}@robots.ox.ac.uk}\\}
\maketitle

\begin{abstract}
Our goal in this paper is the adaptation of image-text models for long video retrieval. Recent works have demonstrated state-of-the-art performance in video retrieval by adopting CLIP, effectively \textit{hitchhiking} on the image-text representation for video tasks. However, there has been limited success in learning temporal aggregation that outperform mean-pooling the image-level representations extracted per frame by CLIP. We find that the simple yet effective baseline of weighted-mean of frame embeddings via query-scoring is a significant improvement above all prior temporal modelling attempts and mean-pooling. In doing so, we provide an improved baseline for others to compare to and demonstrate state-of-the-art performance of this simple baseline on a suite of long video retrieval benchmarks. 

\end{abstract}

\input{01_intro}

\input{02_related}

\input{03_method}
\input{04_exp}
\appendix
\input{appendix}

{\small\bibliographystyle{splncs04}\bibliography{shortstrings,vgg_local,references}}
\end{document}

%% file: 01_intro.tex
\section{Introduction}

Pretrained vision-language models are becoming increasingly ubiquitous due to their impressive performance on a range of downstream tasks with minimal to no additional training data. These models have demonstrated near human-level performance on perception tasks including image classification~\cite{radford2021learning}, image retrieval~\cite{jia2021scaling}, and even object detection~\cite{esmaeilpour2021zero,gu2021open}. A major remaining research question is the successful training and application of vision-language models to tasks requiring higher level cognitive reasoning. One such area, and the focus of this work, is long-form video understanding and its growing body of research~\cite{zhao2019long,wu2019long,wu2022memvit,Wu_2021_CVPR, Han22}.

\begin{figure}
\centering
\begin{subfigure}{.5\textwidth}
    \centering
    \includegraphics[width=.98\linewidth]{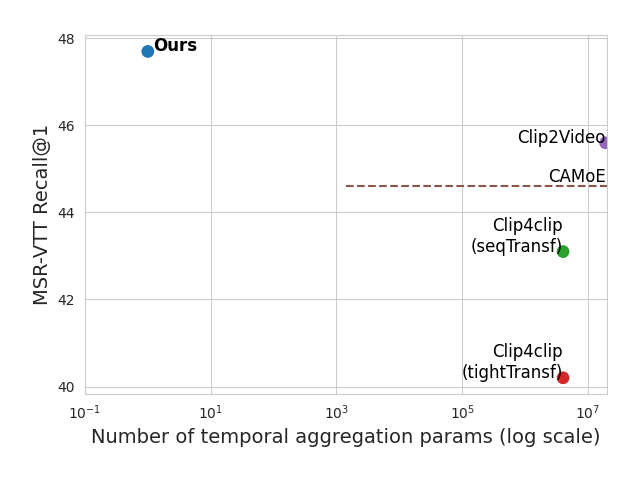}
    \label{fig:cmd_num_frames}
\end{subfigure}%
\begin{subfigure}{.5\textwidth}
    \centering
    \includegraphics[width=.98\linewidth]{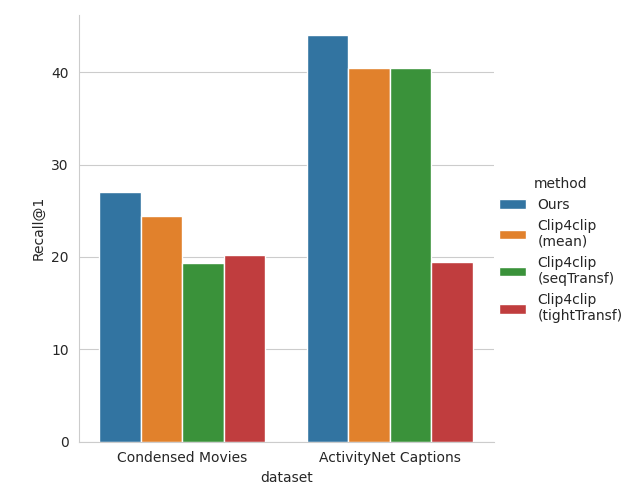}
    \label{fig:cmd_temp_ablation}
\end{subfigure}
\caption{A comparison of performance on text-to-video retrieval between state-of-the-art work initialising with CLIP. Recall@1 performance on MSR-VTT~\cite{xu2016msr} (left) and ActivityNet \& Condensed Movies~\cite{bain2020condensed} (right). Clip2Video~\cite{fang2021clip2video}, CAMoE~\cite{cheng2021improving} and Clip4clip~\cite{Luo2021CLIP4Clip} use involved temporal aggregation methods with many learned parameters. We show that our simple query-scoring baseline (with no learned parameters) outperforms all these works by a large margin. We note the number of parameters for CAMoE is an estimation since there was no public implementation at the time of writing.}
\label{fig:test}
\end{figure}

Machines that can parse long-form videos, understand narrative and abstract concepts, e.g.\ a movie depicting \textit{two friends falling out and then years later making amends}, is a step towards higher level cognitive reasoning. However, progress in this area has been less fruitful when compared to large-scale language models and tasks~\cite{brown2020language,chen2021evaluating}. 
Thus far, vision-language models proven to be effective at analysing short video clips, achieving state of the art when finetuned on tasks such as video classification~\cite{wang2021actionclip,Zhou2021LearningTP,Castro2022FitCLIPRL}, text to video retrieval~\cite{Luo2021CLIP4Clip,Yan2021VideoTextPW,cheng2021improving} and video question answering~\cite{Yang_2021_ICCV,fu2021violet}. 

Whilst recent works have employed pretrained video-text encoders for downstream video tasks~\cite{Bain21,Yan2021VideoTextPW,xu2021videoclip,ge2022bridgeformer}, the current state-of-the-art in text-to-video retrieval employs purely image-text representations, specifically OpenAI's CLIP~\cite{radford2021learning} -- we aptly call this \textit{CLIP-hitchhiking}. This can largely be attributed to the greater scale of image-text datasets~\cite{radford2021learning,jia2021scaling} compared to video-text, by several orders of magnitude. Adapting these models -- originally trained for image data -- to video tasks is still an open question and a growing area of research. Of particular note are tasks involving long-form videos, typically with much smaller amounts of training data, higher degrees of temporal structure and variation between frames. Recent work has proposed learning temporal aggregation layers on top of CLIP representations~\cite{Luo2021CLIP4Clip,cheng2021improving,wang2021actionclip}; however, the performance is comparable or even worse than simply taking the mean of the image representation across all the frames in the video. For the case of long-form videos, with duration of several minutes or more, interesting events might only last a few seconds. Mean-pooling in this case is clearly sub-optimal.

We address this limitation in this paper. Motivated by the fact that long videos can contain many redundant frames, such as a long video of a student studying a math problem, as well as occasionally highly-informative frames, such as a few seconds of footage where said student solves the problem and raises their fist in excitement:  we show that predicting the relevance of each frame and using these scores to perform a simple weighted mean of the frame embeddings outperforms all the more complex temporal modelling attempts and achieves state-of-the-art text-to-video retrieval on ActivityNet Captions~\cite{krishna2017dense}, MSR-VTT~\cite{xu2016msr}, and Condensed Movies~\cite{bain2020condensed}. We investigate three methods for computing the frame relevance scores: (1) Query-scoring, the simplest with no learned parameters, using the frame-level similarity to the text query; (2) Self-attention scoring, a sequence transformer taking frame embeddings as input and outputting scores per frame, conditioned only on video information; and (3) Joint-attention scoring, with the same setup as (2) but additionally with the text query embedding appended to the end of the sequence and thereby conditioning on the query. We demonstrate the improvement of our simple baseline method on using CLIP for long-video classification on the Charades dataset~\cite{sigurdsson2016hollywood}. Our proposed method acts as an improved baseline of mean-pooling for other methods to compare to, especially those which propose aggregation methods on top of CLIP. We further provide insight into the reasons behind the effectiveness of this simple baseline, namely (i) the mean of frame embeddings are mapped to entirely new locations in the embeddings space and (ii) the effect of performance on datasets with differing amounts of data.

%% file: 02_related.tex
\section{Related Work}
We provide a brief overview of the relevant literature on
visual-text representation learning, video-text retrieval, and
long video representation learning. \\

\noindent\textbf{Visual-text representation learning.}
Learning joint visual-text representation learning is a widely studied and growing area of research~\cite{radford2021learning,jia2021scaling,li2021align,wang2016learning,chen2020uniter,li2020oscar,flamingo_deepmind,ALBEF,li2022blip}. Such representations have widespread applications in the real world ranging from semantic video search, zero-shot image classification, and human-robot interaction~\cite{goodwin2021semantically}. Large-scale models trained contrastively on paired visual-text web data has demonstrably shown to learn state-of-the-art image representations capable of impressive zero-shot performance~\cite{radford2021learning}, although these representations are not without their biases~\cite{berg2022prompt,clip_audit}. Works have learned how to leverage noisy speech supervision to learn a better video encoder~\cite{miech2019howto100m}, as well as incorporating self-supervised learning to improve the visual-text representation~\cite{mu2021slip}.
We investigate adopting the large-scale pretrained model CLIP~\cite{radford2021learning} to the domain of videos, particularly those of long-form. \\

\noindent\textbf{Video-text retrieval.}
The text-to-video retrieval area has seen rapid progress over the past few years. First attempts utilised pre-extracted features~\cite{miech18learning}, from classification networks trained for example on ImageNet~\cite{russakovsky2015imagenet} and Kinetics~\cite{Carreira2017}. A large portion of these works investigate how best to aggregate features from these different networks~\cite{Liu19a}, exploring vector quantization~\cite{Wang_2021_CVPR}, incorporating additional modalities such as audio~\cite{Gabeur_2022_WACV,eclipse,everything_at_once}, as well as how to aggregate them over the full video duration (since these features typically have a temporal resolution of a single frame or a couple of seconds).

Most recently, state-of-the-art works have employed CLIP as an image-text backbone and have adapted it to the video setting. Whilst joint text-video pretrained models would be a seemingly better fit, no such models have been made available -- in large part due to the lack of available large scale text-video data. Whilst today's publicly available video-text datasets (Howto100M~\cite{miech2019howto100m}, Youtube8M~\cite{abu2016youtube}, and WebVid10M~\cite{Bain21}) make some headway, this is a far cry from the 400M~\cite{radford2021learning} or 3BN~\cite{zhai2021scaling} diverse image-text pairs today's models are trained on, which far outperform the video-text models trained on less pretraining data. Current state of the art in video tasks do not utilise video-text pretraining effectively whether this is due the infeasible compute required, the marginal gains over image-text pretraining, or even the scale and quality of the visual-text pairs.
\cite{Luo2021CLIP4Clip,fang2021clip2video,cheng2021improving} all propose methods using CLIP with additional temporal modelling on top -- however the temporal modelling performs comparably or worse than taking the mean embedding across the frames~\cite{castro2022fitclip}. A temporal transformer on top of image-level embeddings, has shown demonstrable improvements in other tasks, but typically require vast amounts of training data and cannot be extensively pretrained due to prohibitive cost and lack of large-scale long video-text data. However, \textit{some} temporal modelling must be done since the mean representation can not intelligently aggregate multiple events over a long video.

To overcome this limitation, we focus on the most restricted form of temporal aggregation of frame embeddings: the weighted-mean. In our experimental study, we explore different ways to obtain these weights. \\

\noindent\textbf{Long video representation learning.}
Aggregating temporal information from long videos has been investigated by many works, primarily for video classification~\cite{yue2015beyond,miech17loupe,Gaidon2013TemporalLO,pirsiavash2014parsing,varol18_ltc,adv_pertub,wang2016temporal}. Closest to our work is SCSampler~\cite{Korbar_2019_ICCV} which samples the top-k most salient clips from a long video to use for video classification. SCsampler tackles the sampling with a separate, cheaper model, which is learned separately from model training. The clips selected by the sampling model are then fed to the actual classification network to average scores over the top-k frames. This differs from our work in that it focuses on efficient sampling for video classification by using a cheaper model, learned separately. Our investigation involves weighting frame samples online during training to improve performance and learning.

The joint image-text representation offers a unique advantage to these works where the text-query representation can be used to guide the temporal aggregation. Although more costly, this approach can use the query to guide the temporal aggregation. Works have used query scoring in this way to perform weakly-supervised action and moment localisation, by thresholding frames past a certain similarity with the text query. In a similar vein to these works, we show that using the query to guide the linear weighting of the frame embeddings can achieve state-of-the-art performance without any additional learning.

%% file: 03_method.tex
\section{Temporal Aggregation of Image-Text Representations}
We consider the problem of learning joint text-video representations from a set of video-text pairs $(V, T)$ where $V$ is a video of $K$ frames and $T$ is the corresponding text describing the video.
Specifically, we consider the case where representations are extracted for the text, $T \in \mathbb{R}^d$ and every frame of the video, $V = [I^{(1)}, I^{(2)}, ..., I^{(K)}] \in \mathbb{R}^{K \times d}$, via a pretrained image-text model, such as CLIP~\cite{radford2021learning}. Our goal is to find an aggregation method $\Phi$ that combines the frame representations into a single video-level representation, $\bar{V} = \Phi(V) \in \mathbb{R}^d$, such that semantically similar instances of $\bar{V},T \in \mathbb{R}^d$ are close to each other.

Prior works have instantiated $\Phi$ with self-attention networks~\cite{fang2021clip2video}, squeeze-and-excitation networks~\cite{cheng2021improving}, and even cross-transformer layers with the query~\cite{Luo2021CLIP4Clip}. However, it has been shown that simply taking the mean of every frame embedding achieves comparable or even superior performance to these temporal aggregation attempts on many benchmarks.

This failure of temporal modelling for videos, especially those of long-form is sub-optimal. Video frames have varying degrees of relevance. Motivated by this, as well as the effectiveness of mean-pooling, we propose a straightforward but effective improvement to the uniform mean, inspired by weakly-supervised moment localisation~\cite{mithun2019weakly}, by using query-frame scoring to perform the weighted-mean of frame embeddings. Given a sequence of corresponding per-frame relevance scores, $S = [s_1, s_2, ..., s_K] \in \mathbb{R}^{K}$ where $s_i = I^{(i)} \cdot T $,  we can compute a final embedding for the whole video $\bar{V} \in \mathbb{R}^{d}$ via the weighted-mean.:
\begin{equation}
    \bar{V} = \sum_{k \in K}{w_{k} I}^{(k)} \quad\text{where}\quad w_k = \frac{e^{s_k/\tau}}{\sum_{j \in K}e^{s_j/\tau}},
\end{equation}
where the softmax temperature $\tau$ can be interpreted as a hyperparameter  towards the highest scoring frames. For very small values of $\tau$, this becomes an argmax operation, where the final video embedding is simply the single most relevant frame. Equally, for very large values of $\tau$, the weights become uniform, effectively ignoring the scores. Formally:
\begin{equation}
\lim_{\tau\to 0} \bar{V} \to I^{(k')} \quad\text{where}\quad k' = \operatorname*{argmin}_k S \quad\text{and}\quad \lim_{\tau\to \infty} \bar{V_i} \to \frac{1}{K} \sum_{k \in K}{I^{(k)}}.
\end{equation}
In practice we find the some middle range offers a good balance between weighted relevant frames more as well as capturing the full temporal of content in the video. We explore different values of $\tau$ in our empirical evaluation.

In the following, we describe alternative scoring methods and their complexity (Sec.~\ref{subsec:alternativescoring}), alternative aggregation methods (Sec.~\ref{subsec:alternativeaggregation}), as well as the framing of this problem to video classification (Sec.~\ref{subsec:retrieval-classification}).

\subsection{Alternative Scoring Methods}
\label{subsec:alternativescoring}

\begin{figure}[t]
    \centering
    \includegraphics[width=1\textwidth]{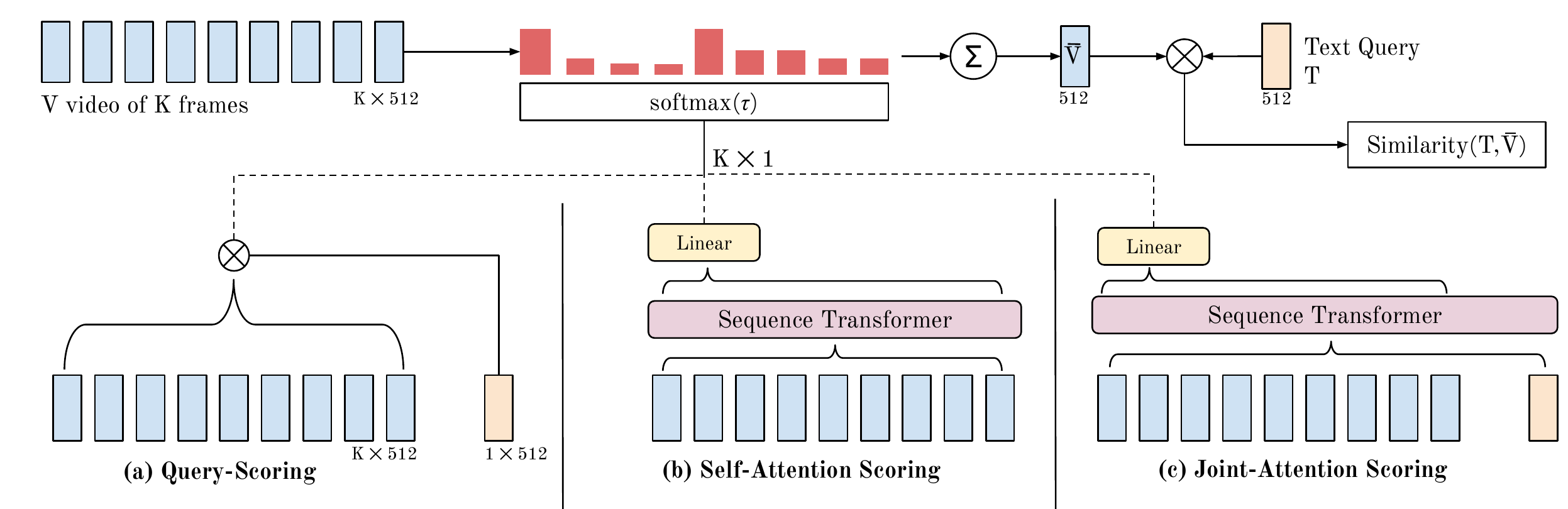}
    \caption{\small{The different scoring methods used to predict the relevance scores of each frame embedding from a video, these are softmaxed as used to compute the weighted-mean single video representation. \textbf{Query-scoring (a)}, the simplest scoring method with no learned parameters, scores each frame by the similarity of each frame embedding with the text query. \textbf{Self-attention scoring (b)}, uses a sequence transformer on the $K \times 512$ frame embeddings, and scores each frame by feed each output frame embedding output sequence through a linear layer $\mathbb{R}^{512} \xrightarrow{} \mathbb{R}^1$. \textbf{Joint-attention scoring (c)}, uses the same approach as (b) with the addition of the text query embedding appended to the input sequence of the transformer.}}
    \label{fig:scoring_methods}
\end{figure}

Whilst the above scoring method is parameter-free, except for the choice of $\tau$, we also investigate more involved methods to predict the scores of each frame. Since the scores can only be used to linearly combine the original frame embeddings, the model is heavily regularised in what it can do and therefore it allows heavy temporal modelling networks to be used but constrains their influence to only linear combinations of the original image-text representation.

\subsubsection{Self-Attention} layers can be used on the frame embeddings to predict the relevance scores, as shown in Figure~\ref{fig:scoring_methods}. The output frame embeddings of the self-attention layers are fed through a linear layer $\mathbb{R}^d \xrightarrow{} \mathbb{R}$ to produce scalar relevance scores per frame. This method has the advantage in that the frame scores $S$ are independent from the query, and therefore also the final video representation $\bar{V}$. This keeps the retrieval complexity to constant time $\mathcal{O}(1)$ (see Table~\ref{tab:complexity}).

\subsubsection{Joint-Attention Scoring} is an extension of the above method, this additionally includes the query as input to the attention layers and performs both cross and self-attention between the query and frame embeddings. Whilst higher in complexity, conditioning the video specific to the query makes sense, as otherwise the definition of frame relevance may be ambiguous.

Neither self-attention layers nor joint-attention layers are new for temporalling modelling, however our instantiation differs in that we do not use the final output embeddings of the attention layers as the video representation, but instead map their output $\mathbb{R}^d \xrightarrow{} \mathbb{R}$ to scalars used to weight the frame embedding averaging.

\subsubsection{Complexity.} The space and time complexities of the different scoring methods are shown in Table~\ref{tab:complexity}, with increasing complexity down the rows. Since query-scoring needs no learnable parameters (like uniform mean), the method can be applied to zero-shot video tasks using image-text only embeddings. The query-conditioned frame aggregation increases retrieval complexity to $\mathcal{O}$, since the weighted-mean is specific to each query -- however in practice we find that 64 or 120 frame embeddings is sufficient for a long video of several minutes. Storing such an array per video is a small increase in space and the dot product operation is marginal. 

The retrieval complexity of query-dependent aggregation can factored down by only employing it for the top $K$ ranked results, and using query-independent aggregation for the full ranking~\cite{Miech2021ThinkingFA}. First a rough ranking is performed on mean embeddings, without query-specific aggregation, and then a more costly query scoring method can be used.

Temporal self-attention and joint attention have the same retrieval and model complexity as prior work temporal modelling attempts -- only with an additional linear layer to map the embeddings to frame scores.

\input{tables/complexity}

\subsection{Alternative Aggregation Methods}
\label{subsec:alternativeaggregation}

\subsubsection{Hard Top-K.}
An alternative to taking the weighted-mean via the softmax scores, can be to take the mean of the hard top-K frames. This is the approach adopted by~\cite{Korbar_2019_ICCV} to select which clips should be aggregated for video classification. Unlike the soft query-scoring which, with the exception of very low values of $\tau$, still includes some amount of information from \textit{every} frame (due to the soft operation), top-k entirely removes them from the aggregation and treats all the top $k$ frames equally.

\subsubsection{Averaging per-frame logits rather than embeddings.}
Similarly, rather than taking the weighted-mean of frame-level embeddings for a single video representation, one can average the similarity logits in order to calculate their similarity to the text. We find that although this performs comparably under zero-shot setting, the hard top-k performance worse when finetuning -- we show results in Section~\ref{subsec:ablation}.

\subsection{Video-to-text retrieval and video classification}
\label{subsec:retrieval-classification}
Whilst the discussed methods have been described in a video-text retrieval setting, it can be equally applied to video classification by formulating the classification task as video-to-text retrieval. The only differences to note would be the complexity analysis, where the video space complexity is no longer a concern since video embeddings need not be stored for text retrieval. Additionally, query-condition aggregation is less of a concern since the number of text queries is fixed to the number to video action labels, which tends to be small.

%% file: tables/complexity.tex
\begin{table}[t]
\centering
\scriptsize
\setlength{\tabcolsep}{11pt}
\caption{$v$ is the number of videos in the retrieval set, $k$ is the number of frames per video, $n$ is the number of layers in the transformer.}
\begin{tabular}{@{}cccc@{}}
\toprule
Scoring Method        & \begin{tabular}[c]{@{}c@{}}Retrieval\\ Complexity\end{tabular} & Model Complexity  & Video Space Complexity \\ \midrule
Mean-pooling            & $\mathcal{O}(1)$                                                                   & $\mathcal{O}(1)$    & $\mathcal{O}(v)$               \\
Query             & $\mathcal{O}(v)$                                                                   & $\mathcal{O}(1)$    & $\mathcal{O}(vk)$                \\
Temporal Self-Attn. & $\mathcal{O}(1)$                                                                   & $\mathcal{O}(n k^2)$ & $\mathcal{O}(v)$     \\
Joint Attn. & $\mathcal{O}(v)$                                                                   & $\mathcal{O}(nk^2)$ & $\mathcal{O}(v)$ \\ \bottomrule
\end{tabular}
\label{tab:complexity}
\end{table}

%% file: 04_exp.tex
\section{Experiments}

In this section, we start by presenting the downstream datasets (Sec.~\ref{subsec:datasets})
and the experiment protocol (Sec.~\ref{subsec:protocol}). Next, we report state-of-the-art results across the chosen suite of long-form video retrieval benchmarks (Sec.~\ref{subsec:results}). Then we perform investigation into the effectiveness of the simple weighted-mean aggregation and compare to alternative methods (Sec.\label{subsec:ablation}).

\subsection{Downstream Datasets}
\label{subsec:datasets}
We now describe the downstream text-to-video retrieval datasets our model is evaluated on, focused on those with long durations, as well as additionally a long video classification dataset.
\\
\noindent\textbf{MSR-VTT~\cite{xu2016msr}} benchmark contains 10K videos from YouTube with 5 captions per video, we trained on 9K videos and report results on the 1K-A~\cite{yu2018joint} test set, this dataset contains the shortest videos of which we evaluate on, averaging 15 seconds.
\\
\noindent\textbf{Condensed Movies Dataset (CMD)~\cite{bain2020condensed}} is a long-form text-video dataset consisting of 34K videos of movie scenes with an average duration of 132 seconds and corresponding high-level semantic textual descriptions. We train, validate and test on the recent challenge split~\cite{CMD_chall} of 32K, 2K, and 1K videos, respectively and compare to the Codalab leaderboard as well as evaluate results on other competing methods. Within a long movie scene there is a large variation among the events that occur and the information between shots, aggregating this over several minutes to match to a high-level semantic description requires a high degree of video understanding.
\\
\noindent\textbf{ActivityNet Captions~\cite{krishna2017dense}} consists of 20K videos from YouTube focused primarily on actions, annotated with 100K sentences. The training set consists of 10K videos, and we use the val1 set of videos to report results. With an average video duration of 180 seconds, these are the longest videos we evaluate on -- capturing a diverse range of events and actions within the two minutes. The captions consist of a sequence of descriptions of localised moments within the video. We employ the standard paragraph-to-video retrieval~\cite{Bain21} protocol when training and testing by concatenating the text sequences and evaluating on the whole long-form video.

\noindent\textbf{Charades~\cite{sigurdsson2016hollywood}} is a video classification dataset consisting of daily activities with an average duration of 30 seconds. The classification is multilabel and multiclass in that a video can contain multiple different actions at different times. This is a valuable setting for long-form video understanding since the action classes vary over the duration.

\subsection{Experiment Protocol}
\label{subsec:protocol}
We use CLIP ViT-B/16~\cite{radford2021learning} in all experiments and finetune the model end-to-end with the Adam optimizer~\cite{DBLP:journals/corr/KingmaB14} (learning rate set to 5e-7). Finetuning is done on two GPUs with a batch size of eight per GPU, and sample 16 frames from each video during training, randomly sampled within the video clip (we found this superior to the sub-segment sampling in~\cite{Bain21}). At test time 120 frames are sampled uniformly from the video, irrespective of their length, unless specified otherwise. For the text, words are dropped randomly during training with a probability of 10\%. For query-scoring, we use the $\tau=0.1$ across all datasets to demonstrate the consistent performance (although optimal values might vary slightly between datasets). For the self and joint attention scoring methods, we use single-layer networks as was minimal difference in performance when using additional layers.
\subsection{Results}
\label{subsec:results}
\subsubsection{Comparison to the state of the art.}
We present the text-to-video retrieval results of the query-scoring method on MSR-VTT, ActivityNet and CondensedMovies in Tables~\ref{tab:msr-sota} and~\ref{tab:cmd-sota}. We achieve state-of-the art performance on all three datasets, significantly outperforming prior work aggregation methods using a CLIP backbone with millions of learned parameters. In contrast, our query-scoring method has only a single parameter. These results demonstrate the surprising effectiveness of \textit{weighted-mean} embeddings, and the limitations of current, more involved temporal aggregation methods. Query-scoring acts an improved baseline to proposed temporal aggregation methods.

\input{tables/msrvtt}
\input{tables/cmd}

We also compare to results on the long video classification dataset Charades in Tab.~\ref{tab:charades}. CLIP with query-scoring achieves the same performance as ActionCLIP~\cite{wang2021actionclip} which uses temporal modelling, averages predictions over 320 total frames, and a set of prompt templates. Query-scoring uses none of these yet offers similar performance, and outstanding improvements to CLIP4CLIP \textit{seqTransf} and the baseline of mean pooling the frame embeddings.

\noindent\textbf{Test-time normalisation}. Recent concurrent works achieve state of the art performance via CLIP with test-time normalisation such as QueryBank~\cite{cheng2021improving,bogolin2022cross} and dual softmax normalisation. The latter however requires access to all queries at test-time, which is not appropriate for real-world tasks. Both of these can be added to our method to achieve superior performance, but is not the focus of this work since it is separate to temporal modelling.

\input{tables/charades}

\subsection{Ablation study}
\label{subsec:ablation}

\noindent\textbf{Alternative scoring methods.}
In Table~\ref{tab:attn_compare} we show the performance of alternative scoring methods, and that the performance is notably high across the board -- even without query information. This indicates that the strong baseline of weighted-mean of image-level CLIP embeddings is the best current approach of temporal modelling. The seemingly consistent boost no matter the scoring method indicates that the benefit is mainly afforded by restricting the aggregation $\Phi$ to linear weighted of the image embeddings. For ActivityNet, temporal self-attention scoring performs significantly better than the other scoring methods which is in contrast to CMD and MSR-VTT. One possible explanation for this is that ActivityNet captions are long paragraphs containing dense descriptions on the video by concatenating localised descriptions. Such a dense video description would be less useful when query-scoring since most frames relate to the query. This is unlike Charades where an action class might only correspond to a few seconds of a 30-second video amongst a sequence of other actions. Instead, the dense description setting of ActivityNet becomes a case of removing noisy frames, which we believe can be done without query-level information.

In contrast CMD, movies with long videos but extremely concise descriptions pertaining to a sequence of a few events in the video, affords the greatest benefits.

\input{tables/attention_compare}

\input{tables/scoring_compare}
\noindent\textbf{Alternative aggregation methods.} Comparing different aggregation methods in Table~\ref{tab:scoring_compare}, we find that averaging over features is considerably better in both zero-shot and finetuning settings. Hard top-K seems to offer a comparable improvement over the baseline in the zero-shot setting, however we find soft scoring features for the finetune setting is most optimal. A smoother feature selection of frames seems to be more favourable for training. 

\begin{figure}[t]
\centering
\begin{subfigure}{.5\textwidth}
    \centering
    \includegraphics[width=.98\linewidth]{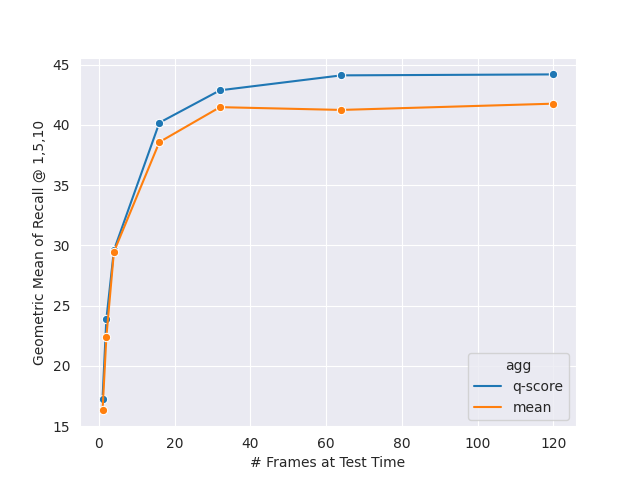}
    \label{fig:cmd_num_frames}
\end{subfigure}%
\begin{subfigure}{.5\textwidth}
    \centering
    \includegraphics[width=.98\linewidth]{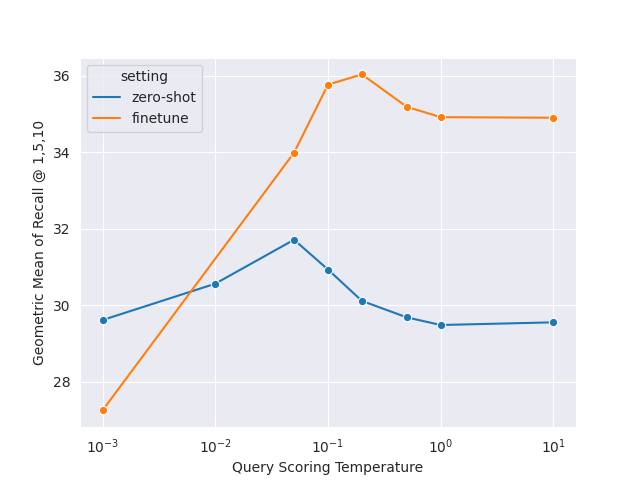}
    \label{fig:cmd_temp_ablation}
\end{subfigure}
\caption{Downstream performance effects of varying the number of frames at test time (left) and the query-scoring temperature $\tau$ (right), on the CMD test set for text-to-video retrieval.}
\label{fig:test}
\end{figure}

\noindent\textbf{Effect of number of frames.}
Unsurprisingly performance of both the baseline and query-scoring improves with increasing the number of input frames at test time, albeit with diminishing after ~1fps, shown in Figure~\ref{fig:test} (left). The improvement of query-scoring over the baseline also increases with the number of frames. Intuitively this makes sense, more input frames means there are more relevant frames to pick from and less relevant frames to ignore. This further motivates the use case of frame relevance scoring for long-form videos.

\noindent\textbf{Effect of scoring temperature.}
We find a temperature range between 0.05 and 0.15 is consistently optimal across long range datasets and tasks. Figure~\ref{test} (right) shows the effect of temperature on the CMD test set under both zero-shot and finetuning conditions. Zero-shot performance tends to be slightly better with smaller values of $\tau$, which indicates the effect of $\tau$ during the learning process. For extremely large values of $\tau$, gradients only pass from the most similar text query which might cause drifting errors in the learning process (for the case where the initial similarity is wrong). \\

\noindent\textbf{Why are weighted-mean frame embeddings so effective?} \\

\noindent\textbf{a) Insufficient training data for learning new long-video text representations.} We find the relative performance boost of mean-weighted embeddings compared to more complex and learned temporal aggregations is reduced with the larger scale downstream datasets. This implies that with enough long video-text pairs, the more complex modelling attempts can outperform this simple baseline.

\noindent\textbf{b)The mean of frame embeddings captures distinct information.}
Given that CLIP embeddings are trained for the single image-text setting, it is surprising that taking the mean over many frames with vastly different content still performs well. For example, it is possible that the mean of embeddings from two semantically different frames maps to a semantically incorrect new space in the embeddings. To investigate whether this happens, we train both a linear classifier and a multi-layer perceptron (MLP) to classify between single-frame embeddings and mean embeddings from 16 frames sampled from a long video in CMD. We find that both are able to easily classify between these two, even with zero-shot embeddings (Table~\ref{tab:vmean_class}). These results suggest the mean-frame embeddings are mapped to entirely new locations in the embedding space, disjoint from the single-frame embeddings. This is encouraging since it suggests that CLIP can learn to capture multi-frame information within the 512 embeddings -- and hence the strong baseline out weighted-mean performing so well.

\input{tables/n_frame_classification}

\noindent\textbf{c) Query scoring during training improves single-frame representation.} The performance boost of query scoring after finetuning could be attributed to either (i) test time improvements by ignoring irrelevant improvements and/or (ii) improvements to the image-text level representation during training by contrastively learning on more semantically relevant frames. In order to investigate whether (ii) is true we evaluate on CMD test set retrieval in the 1-frame setting, to evaluate the single frame representation (Table~\ref{tab:vmean_class}). We find that query-scoring performance provides notable improvement to the image-text representation, indicating that such a method is valuable during the video-text learning process over the baseline of mean-pooling.
\input{tables/single_frame_eval}

\subsubsection{Are the frame scores semantically meaningful?} 
\begin{figure}[!t]
\centering
\begin{subfigure}{0.95\textwidth}
    \centering
    \includegraphics[width=.98\linewidth]{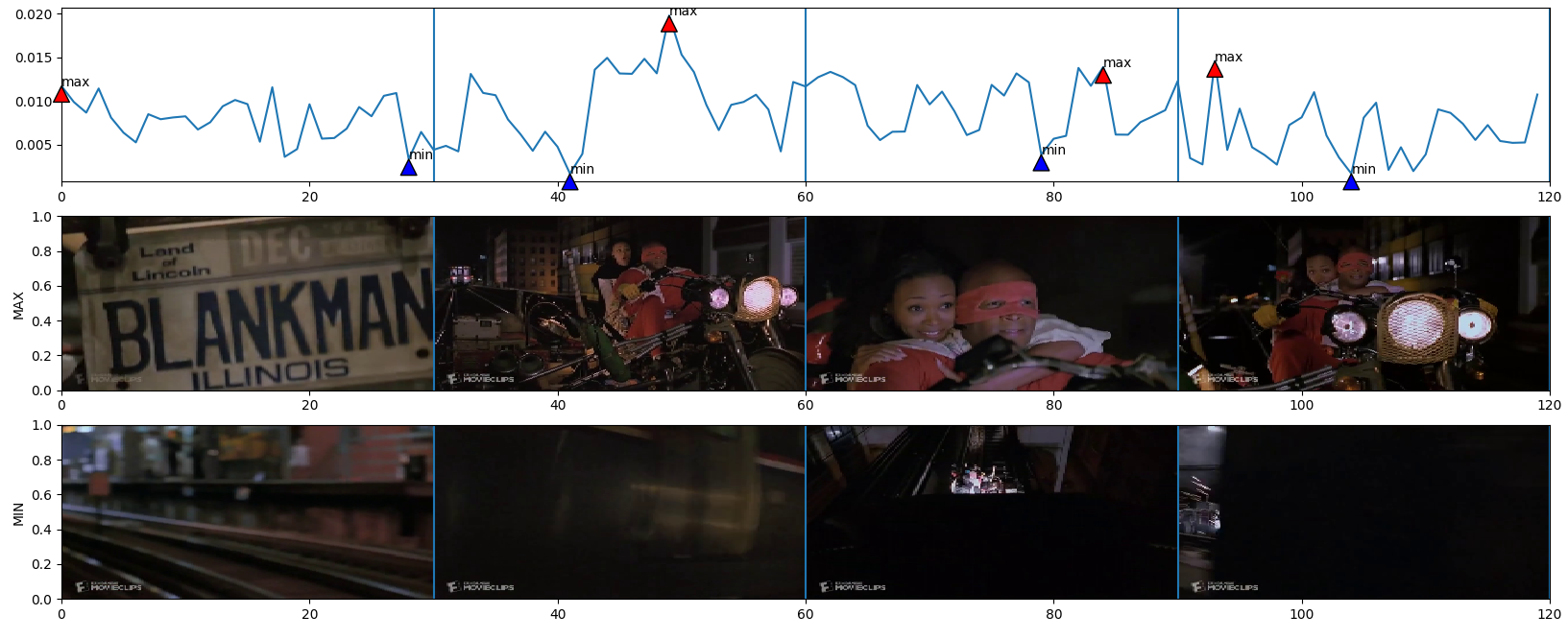}
    \label{fig:motorbike}
    \caption{``Blankman gives Kimberly a ride on his wacky motorcycle.''}
\end{subfigure}\\
\begin{subfigure}{0.95\textwidth}
    \centering
    \includegraphics[width=.98\linewidth]{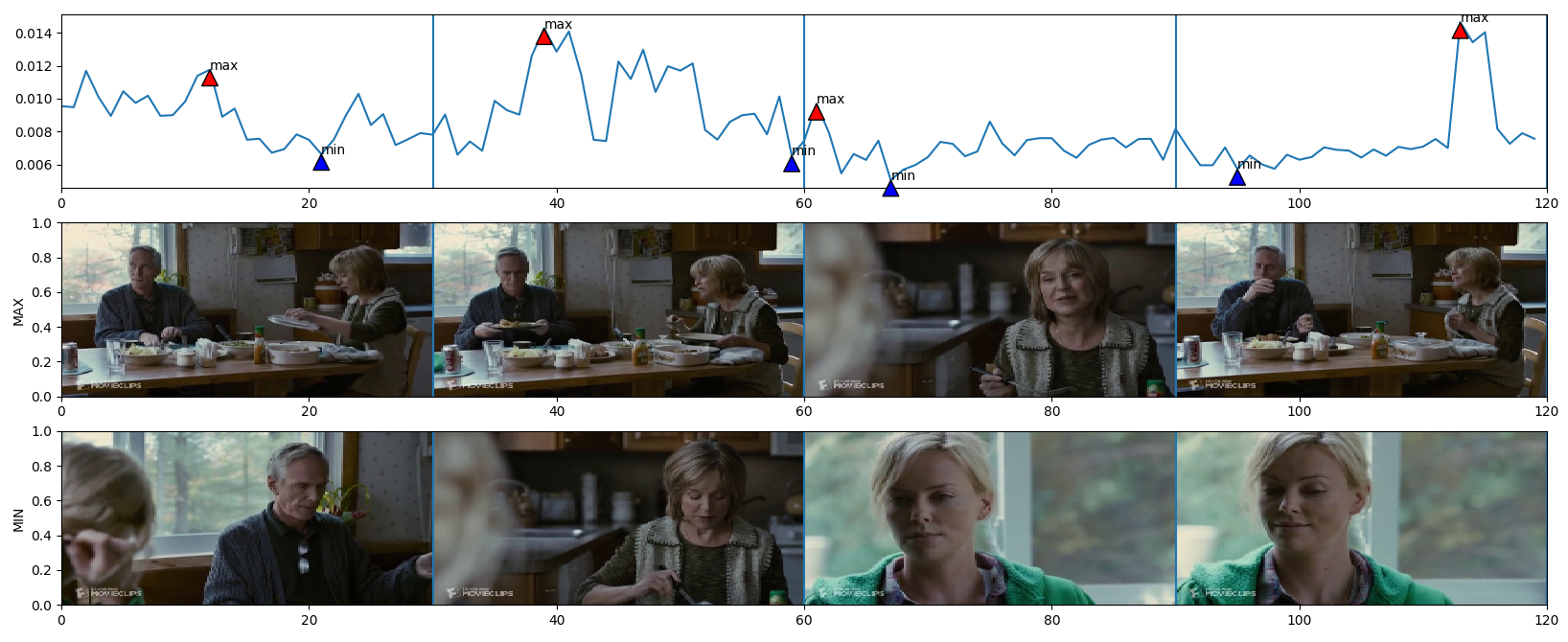}
    \label{fig:parents}
    \caption{``Mavis has an awkward lunch with her parents.''}
\end{subfigure}
\caption{\small{Visualisation of query-scoring from unseen videos in the Condensed Movies test set and their corresponding textual descriptions. The first row shows the weights assigned to each frame, with the maximum and minimum scores per segment marked and their corresponding frames in the middle and bottom rows respectively. The highest scoring frames in both (a) and (b) have high relevance to the text query and contain more information than the lowest scoring frames containing background frames or single person close-up shots.}}
\label{fig:query_scoring_qual}
\vspace{-0.8em}
\end{figure}

The notable improvement gains via frame scores implies some semantic relevance of the higher scoring frames. To investigate this we show qualitative results of the query scoring for CMD unseen videos. In Figure~\ref{fig:query_scoring_qual} we see that highest scoring frames are those with semantic similarity to the test, for example the frames containing the motorbike as well as the license registration of the character name in the title. Additionally, we see that the lowest scoring frames contain less information and have less relevance to the query. By utilising frame scoring during training, the constrastive loss is weighted less towards these irrelevant frames which could otherwise harm the representation.

\section{Conclusion}
\vspace{-0.8em}
To conclude, we propose three simple ways to mean-weight frame embeddings from a joint image-text representation for long video retrieval and classification -- with and without query information, in doing so picking out the most salient frames. Our method provides a strong baseline, outperforming all prior works across four datasets, including attempts at more complicated temporal modelling. Our experiments uncover some insight into the benefits afforded by this highly constrained temporal aggregation and the challenges posed for more involved temporal modelling. Future work could look into tackling the lack of large-scale data needed to effectively learn effective long-form video representations. This could be done by employing self-supervised learning in addition to the scarcer textual supervision for long-form video data.

%% file: tables/msrvtt.tex
\begin{table}[t]
\setlength{\tabcolsep}{5pt}
\scriptsize
\caption{\small{Comparison to state-of-the-art results on MSR-VTT 1k-A for text-to-video retrieval. The bottom section compares to methods using CLIP as the backbone image-text encoder, their different temporal aggregation methods (agg.), and the number of parameters learned for the aggregation. ``$-$'' indicates an unknown value either due to no official public implementation or lack of reporting in the paper. \textbf{R@k:} Recall@K,\textbf{MedR:} Median Rank, \textbf{MnR:} Mean Rank.}}
\centering
\begin{tabular}{@{}rccrrrrr@{}}
\toprule
\multicolumn{3}{c}{\textbf{Method}}        & \multicolumn{1}{c}{\textbf{R@1$\uparrow$}} & \multicolumn{1}{c}{\textbf{R@5$\uparrow$}} & \multicolumn{1}{c}{\textbf{R@10$\uparrow$}} & \multicolumn{1}{c}{\textbf{MedR$\downarrow$}} & \multicolumn{1}{c}{\textbf{MnR$\downarrow$}} \\ \midrule
\multicolumn{3}{c}{JSFusion~\cite{yu2018joint}}                                                                    & 10.2                             & 31.2                             & 43.2.4                              & 12                                &   -                               \\
\multicolumn{3}{c}{CE~\cite{Liu19a}}                                             & 20.9                             & 48.8                             & 62.4                              & 6                                 &   -                               \\
\multicolumn{3}{c}{MMT~\cite{gabeur2020multi}}                                                                    & 26.6                             & 57.1                             & 69.6                              & 4                                 &    -                              \\
\multicolumn{3}{c}{SSB~\cite{patrick2020support}}                                                                     & 30.1                             & 58.5                             & 69.3                              & 3                                 &  -                                \\
\multicolumn{3}{c}{TeachText~\cite{Croitoru21a}}                                                                     & 29.6                             & 61.6                             & 74.2                              & 3                                 &   -                               \\
 \multicolumn{3}{c}{Frozen~\cite{Bain21}}                                                                                  & 32.5                             & 61.5                             & 71.2                              & 3                                 &     -                             \\ \hline
\textbf{Method}  & \textbf{agg.}        & \textbf{\#agg. params} & &  &  &  &  \\ \midrule
{Clip4clip~\cite{Luo2021CLIP4Clip}}    & mean                 & 0                                                                   & 43.1                             & 70.4                             & 80.8                              & 2                                 & 16.2                             \\
{Clip4clip~\cite{Luo2021CLIP4Clip}}    & tightTransf          & 4M                                                                 & 40.2                             & 71.5                             & 70.5                              & 2                                 & 13.4                             \\
  {Clip4clip~\cite{Luo2021CLIP4Clip}}  & seqTransf            & 4M                                                                 & 44.5                             & 71.4                             & 81.6                              & 2                                 & 15.3                             \\
CAMoE~\cite{cheng2021improving}     & S.E attn.            & -                                                                 & 44.6                             & 72.6                             & 81.8                              & 2                                 & 13.3                             \\
Clip2Video~\cite{fang2021clip2video}   & TDB,TAB          & 19M                                                                & 45.6                             & 72.6                             & 81.7                              & 2                                 & 14.6                             \\ \midrule
\multicolumn{1}{c}{Ours}                     & Q-score              & 1                                     & \textbf{47.7}                    & \textbf{74.1}                    & \textbf{82.9}                     & 2                                 & \textbf{11.5}                    \\ \bottomrule
\end{tabular}
\label{tab:msr-sota}
\end{table}

%% file: tables/cmd.tex
\begin{table}[t]
\centering
\scriptsize
\setlength{\tabcolsep}{3pt}
\caption{\small{Comparison to the start-of-the-art results on Condensed Movies and ActivityNet Challenge for text-to-video retrieval.}}
\begin{tabular}{@{}crrrrr|rrrrr@{}}
\toprule
                      & \multicolumn{5}{c|}{\textbf{Condensed Movies}}                                                                                                                                 & \multicolumn{5}{c}{\textbf{ActivityNet}}                                                                                                                                      \\
\textbf{Method}       & \multicolumn{1}{c}{\textbf{R@1}} & \multicolumn{1}{c}{\textbf{R@5}} & \multicolumn{1}{c}{\textbf{R@10}} & \multicolumn{1}{c}{\textbf{MdR}} & \multicolumn{1}{c|}{\textbf{MnR}} & \multicolumn{1}{c}{\textbf{R@1}} & \multicolumn{1}{c}{\textbf{R@5}} & \multicolumn{1}{c}{\textbf{R@10}} & \multicolumn{1}{c}{\textbf{MdR}} & \multicolumn{1}{c}{\textbf{MnR}} \\ \midrule
TeachText             & 12.1                             & 27.4                             & 37.5                              & -       & -                                 & 25.0                             & 58.7                             & -                                 & 4                              & -                                \\
Frozen                &  12.6                                &   28.4                               &  36.3                       &  25                      &   -   & 28.8                    & \multicolumn{1}{c}{-}   & 60.9                              & 3                              & -                                \\
Clip4clip (mean)      & 24.4                             & 48.2                             & 58.2                              & 6                              & 46.2                              & 40.5                             & 72.4                             & -                                 & 2                              & 7.4                              \\
Clip4clip (seqTransf) & 19.3                             & 44.4                             & 55.3                              & 9                              & 55.2                              & 40.5                             & 72.4                             & -                                 & 2                              & 7.5                              \\ \midrule
Ours (q-score)        & \textbf{27.0}                    & \textbf{52.3}                    & \textbf{61.2}                     & \textbf{5}                     & \textbf{41.2}                     & \textbf{44.0}                    & \textbf{74.9}                    & \textbf{86.1}                     & 2                              & \textbf{5.8}                     \\ \bottomrule
\end{tabular}
\label{tab:cmd-sota}
\end{table}

%% file: tables/charades.tex
\begin{table}[t]
\setlength{\tabcolsep}{14pt}
\scriptsize
\centering
\caption{Multi-label classification results on the Charades dataset, where mAP is mean average precision. ActionCLIP results are computed by average 32 frame predictions over 10 (spatial) by 3 (temporal) views.}
\begin{tabular}{@{}clllr@{}}
\toprule
\multicolumn{1}{c}{Backbone}                                                                   & Aggregation         & Frame      & Finetune         & mAP  \\ \midrule

\multirow{4}{*}{\begin{tabular}[c]{@{}c@{}}CLIP\\ (ViT-B/16)\end{tabular}}                     & ActionCLIP~\cite{wang2021actionclip}          & 32$_{\times10\times3}$ & \cmark & 44.6 \\
                                                                                               & Clip4clip$_{seqTr}$ & 32 & \cmark & 32.0      \\ \cline{2-5}
                                                                                               & Mean           & 32 & \cmark & 33.0   \\
                                                                                               & Q-scoring           & 32         & \cmark & \textbf{44.9} \\
                                                                                               & Temp. S-A           & 32         & \cmark & 36.3 \\
                                                                                               & Joint. S-A           & 32         & \cmark & 42.2 \\
                                                                                               
                                                                                               \midrule
\multirow{2}{*}{\begin{tabular}[c]{@{}c@{}}CLIP\\ (ViT-B/16)\end{tabular}} & Mean                & 32         & \xmark           & 17.5 \\
                                                                         & Q-scoring           & 32         & \xmark          & \textbf{21.1} \\
\bottomrule
\end{tabular}
\label{tab:charades}
\end{table}

%% file: tables/attention_compare.tex
\begin{table}[t]
\centering
\caption{Comparison of the different proposed scoring methods on MSR-VTT, ActivityNet Captions and Condensed Movies test sets for text-to-video retrieval..}
\scriptsize
\begin{tabular}{@{}r|rrrr|rrrr|rrrr@{}}
\toprule
\multicolumn{1}{c|}{\multirow{2}{*}{\textbf{\begin{tabular}[c]{@{}c@{}}Scoring\\ Method\end{tabular}}}} & \multicolumn{4}{c|}{\textbf{MSR-VTT}}                                                                                                       & \multicolumn{4}{c|}{\textbf{Condensed Movies}}                                                                                              & \multicolumn{4}{c}{\textbf{ActivityNet Captions}}                                                                                          \\
\multicolumn{1}{c|}{}                                                                                   & \multicolumn{1}{c}{\textbf{R@1}} & \multicolumn{1}{c}{\textbf{R@5}} & \multicolumn{1}{c}{\textbf{R@10}} & \multicolumn{1}{c|}{\textbf{MnR}} & \multicolumn{1}{c}{\textbf{R@1}} & \multicolumn{1}{c}{\textbf{R@5}} & \multicolumn{1}{c}{\textbf{R@10}} & \multicolumn{1}{c|}{\textbf{MnR}} & \multicolumn{1}{c}{\textbf{R@1}} & \multicolumn{1}{c}{\textbf{R@5}} & \multicolumn{1}{c}{\textbf{R@10}} & \multicolumn{1}{c}{\textbf{MnR}} \\ \midrule
Baseline                                                                                                & 44.4                             & 71.6                             & 79.8                              & 12.8                              & 24.4                             & 48.2                             & 58.2                              & 46.2                              & 42.0                             & 73.1                             & 84.6                              & 7.4                              \\ \hline
Query                                                                                                   & \textbf{47.7}                    & \textbf{74.1}                    & 82.9                              & 11.5                              & \textbf{27.0}                    & \textbf{52.3}                    & \textbf{61.2}                     & \textbf{41.2}                     & 44.0                             & 74.9                             & 86.1                              & 5.8                              \\
Temporal Self-Attn.                                                                                     & 46.2                             & 71.4                             & 81.6                              & 12.9                              & 26.2                             & 51.9                             & 62.4                              & 41.5                              & \textbf{44.9}                    & \textbf{75.9}                    & \textbf{86.9}                     & \textbf{5.6}                     \\
Joint Attn.                                                                                             & 45.7                             & 73.8                             & \textbf{83.6}                     & \textbf{10.6}                     & 26.4                             & 51.1                             & 62.0                              & 43.0                              & 43.1                             & 74.1                             & 85.5                              & 6.7                              \\ \bottomrule
\end{tabular}
\label{tab:attn_compare}
\end{table}

%% file: tables/scoring_compare.tex
\begin{table}[t]
\centering
\scriptsize
\setlength{\tabcolsep}{9pt}
\caption{\small{Comparison of different frame aggregation methods and aggregation sources and their respective performance on zero-shot Condensed Movies, zero-shot MSR-VTT, and finetuned MSR-VTT text-to-video retrieval settings denoted by CMD ZS, MSR ZS and MSR FT repsectively. The reported value is the geometric mean of R@\{1,5,10\} to text-to-video retrieval performance. HParam denotes the hyperparameter selection for the chosen method. $\dagger$ Since training , training with $K=60$ is not possible, so this setting is trained on $K=8$ and evaluated on $K=60$.}}
\begin{tabular}{@{}ccrrrrrr@{}}
\toprule
\multicolumn{1}{l}{}                                                               & \multicolumn{1}{l}{}            & \multicolumn{2}{c}{\textbf{CMD ZS}}                     & \multicolumn{2}{c}{\textbf{MSR ZS}}                     & \multicolumn{2}{c}{\textbf{MSR FT}}                     \\ \midrule
\multirow{2}{*}{\textbf{\begin{tabular}[c]{@{}c@{}}Aggregation\\ Method\end{tabular}}} & \multirow{2}{*}{\textbf{HParam}} & \multicolumn{2}{c}{Agg. source}                         & \multicolumn{2}{c}{Agg. source}                         & \multicolumn{2}{c}{Agg. source}                         \\ \cmidrule(l){3-8} 
                                                                                   &                                 & \multicolumn{1}{c}{Score} & \multicolumn{1}{c}{Feature} & \multicolumn{1}{c}{Score} & \multicolumn{1}{c}{Feature} & \multicolumn{1}{c}{Score} & \multicolumn{1}{c}{Feature} \\ \midrule
Mean-pooling                                                         &                                   &  27.6                  &  29.5                               & 48.3                             &  49.5                       & 60.5                           & 62.2                             \\ \hline
\multirow{4}{*}{Top-K}                                          & K=1                         & 28.8                  & 28.8                        &  48.8                       & 48.8                        & 63.3                          & 63.4                            \\
                                                                                   & K=8                       & 32.6                   & 33.3                        &  50.9                       & 50.3                          & 63.3                          & 64.5                            \\                    
                                                                                   & K=60$\dagger$                       & 30.0                   & 30.0                        &  51.2                     &  50.2                        &  61.9                         &  64.0                           \\ \midrule
\multirow{3}{*}{Query-scoring}                           & $\tau$=0.01                    &  27.1                  & 30.6                        & 48.9                          & 49.0                     & 62.4                           & 64.4                            \\
                                                                                   & $\tau$=0.1                      &  28.8                 & 30.9                        & 50.7                          & 50.5                      &  \textbf{63.8}                         & \textbf{65.4}                           \\
                                                                                   & $\tau$=1.0                      &  27.7                 & 29.5                        &  48.5                        &  49.6                      & 60.1                          &  62.8                           \\ \bottomrule
\end{tabular}
\label{tab:scoring_compare}
\end{table}

%% file: tables/n_frame_classification.tex
\begin{table}[t]
\centering
\setlength{\tabcolsep}{14pt}
\caption{Classification accuracy between normalised single-frame embeddings and 16-frame mean-pooled embeddings from CMD (training on the train set and testing on the test set). A single linear layer with high accuracy can discriminate between single-frame and 16-frame embeddings.}
\scriptsize
\begin{tabular}{@{}lll@{}}
\toprule
\textbf{}  & \multicolumn{2}{c}{Test accuracy (\%)} \\ \cmidrule(l){2-3} 
Classifier & Zero-Shot         & Finetuned        \\ \midrule
Linear     & 89.0                   & 90.4               \\
MLP        & 97.4              & 98.2               \\ \bottomrule
\end{tabular}
\label{tab:vmean_class}
\end{table}

%% file: tables/single_frame_eval.tex
\begin{table}
\centering
\scriptsize
\setlength{\tabcolsep}{8pt}
\vspace{-0.5em}
\caption{Single-frame retrieval results on CMD test set. Finetuning with query scoring improves the image-level representation -- showing the effectiveness of constrastive learning on video-text pairs with relevance scoring on frames.}
\begin{tabular}{@{}crrrrr@{}}
\toprule
\textbf{Scoring Method} & \textbf{R@1}  & \textbf{R@5}  & \textbf{R@10} & \textbf{MedR} & \textbf{MnR} \\ \midrule
Mean  & 8.6          & 19.6       & 25.8        & 56.0  & 151 \\
Q-scoring & 9.0        & 21.1        & 27.3          & 52.5 &  148\\ \bottomrule
\end{tabular}
\vspace{-0.5em}
\label{tab:single-frame}
\end{table}

%% file: appendix.tex
\section*{Appendix} 

\title{Appendix}
\section{MSR-VTT Full Split}

In Table~\ref{tab:msr-sota-full} We additionally report results on the MSR-VTT full split which consists of 6513, 497 and 2990 videos for training, validation and testing respectively. Query-scoring outperforms all prior work, except in mean rank, which could be attributed to the smaller amounts of training data.
\input{tables/msrvtt_full}
\section{Extended Comparison of Aggregation Methods}

Finetuning on Condensed Movies also shows a pronounced boost for query-scoring when compared to alternative aggregation methods: hard top-k and mean-pooling (see Table~\ref{tab:score_compare_cmd}). Further demonstrating the benefit of query-scoring in the finetuning setting.
\input{tables/scoring_compare_cmd}

%% file: tables/msrvtt_full.tex
\begin{table}[h!]
\setlength{\tabcolsep}{5pt}
\vspace{-2em}
\scriptsize
\caption{\small{Comparison to state-of-the-art results on MSR-VTT full split with 7k training for text-to-video retrieval. The bottom section compares to methods using CLIP as the backbone image-text encoder, their different temporal aggregation methods (agg.), and the number of parameters learned for the aggregation. ``$-$'' indicates an unknown value either due to no official public implementation or lack of reporting in the paper.}}
\centering
\begin{tabular}{@{}rccrrrrr@{}}
\toprule
\multicolumn{3}{c}{\textbf{Method}}        & \multicolumn{1}{c}{\textbf{R@1$\uparrow$}} & \multicolumn{1}{c}{\textbf{R@5$\uparrow$}} & \multicolumn{1}{c}{\textbf{R@10$\uparrow$}} & \multicolumn{1}{c}{\textbf{MedR$\downarrow$}} & \multicolumn{1}{c}{\textbf{MnR$\downarrow$}} \\ \midrule
\multicolumn{3}{c}{JSFusion~\cite{yu2018joint}}                                                                    & 10.2                             & 31.2                             & 43.2.4                              & 12                                &   -                               \\
\multicolumn{3}{c}{CE~\cite{Liu19a}}                                             & 10.0                             & 29.0                             & 41.2                              & 16                                 &   86.8                               \\
\multicolumn{3}{c}{TeachText~\cite{Croitoru21a}}                                                                     & 15.0                             & 38.5                             & 51.7                              & 10                                 &   -                               \\ \hline
\textbf{Method}  & \textbf{agg.}        & \textbf{\#agg. params} & &  &  &  &  \\ \midrule

Clip2Video~\cite{fang2021clip2video}   & TDB,TAB          & 19M                                                                & 29.8                             & 55.5                             & 66.2                              & 4                                 & 45.4                             \\
CAMoE~\cite{cheng2021improving}     & S.E attn.            & -                                                                 & 32.9                             & 58.3                             & 68.4                              & 3                                 & 42.6                             \\  \midrule

\multicolumn{1}{c}{Ours}    & Q-score  &    1 & 34.9 & 59.4 & 68.7  & 3  & 49.4                     \\

\bottomrule
\end{tabular}
\label{tab:msr-sota-full}
\vspace{-2em}
\end{table}

%% file: tables/scoring_compare_cmd.tex
\begin{table}[h!]
\centering
\vspace{-1em}
\caption{\scriptsize{Comparison between different scoring methods when finetuning on Condensed Movies, results are shown for text-to-video retrieval.}}
\scriptsize
\setlength{\tabcolsep}{8pt}
\begin{tabular}{@{}ccrrrrr@{}}
\toprule
\textbf{Aggregation Method }            & \textbf{Hparam}      & \multicolumn{1}{c}{\textbf{R@1$\uparrow$}} & \multicolumn{1}{c}{\textbf{R@5$\uparrow$}} & \multicolumn{1}{c}{\textbf{R@10$\uparrow$}} & \multicolumn{1}{c}{\textbf{MedR$\downarrow$}} & \multicolumn{1}{c}{\textbf{MnR$\downarrow$}}  \\ \midrule
Mean-pooling                   &              & 25.6      &  49.2    &   59.9   & 6     & 42.2   \\ \hline
\multirow{3}{*}{Top-K}         & K=1         & 19.6 & 43.8 & 52.9 & 8    & 54.0 \\
                               & K=8         & 25.4 & 50.2 & 58.1 & 5    & 51.6 \\
                               & K=60        & 26.0 & 51.6 &  61.1 & 6   & 41.6     \\ \hline
\multirow{3}{*}{Query-scoring} & $\tau$=0.01 & 24.0 & 47.4 & 57.8 & 6    & 52.8 \\
                               & $\tau$=0.1  & \textbf{27.3} & \textbf{52.8} & \textbf{62.0} & \textbf{4}    & \textbf{40.4} \\
                               & $\tau$=1.0  & 25.5 & 49.5 & 60.3 & 6    & 42.3 \\ \bottomrule
\end{tabular}
\label{tab:score_compare_cmd}
\vspace{-1em}
\end{table}

%% file: main.bbl
\begin{thebibliography}{10}
\providecommand{\url}[1]{\texttt{#1}}
\providecommand{\urlprefix}{URL }
\providecommand{\doi}[1]{https://doi.org/#1}

\bibitem{CMD_chall}
Condensed movies challenge.
  \url{https://www.robots.ox.ac.uk/~vgg/research/condensed-movies/challenge.html},
  accessed: 2022-03-06

\bibitem{abu2016youtube}
Abu-El-Haija, S., Kothari, N., Lee, J., Natsev, P., Toderici, G., Varadarajan,
  B., Vijayanarasimhan, S.: Youtube-8m: A large-scale video classification
  benchmark. arXiv preprint arXiv:1609.08675  (2016)

\bibitem{clip_audit}
Agarwal, S., Krueger, G., Clark, J., Radford, A., Kim, J.W., Brundage, M.:
  Evaluating clip: towards characterization of broader capabilities and
  downstream implications. arXiv preprint arXiv:2108.02818  (2021)

\bibitem{flamingo_deepmind}
Alayrac, J.B., Donahue, J., Luc, P., Miech, A., Barr, I., Hasson, Y., Lenc, K.,
  Mensch, A., Millican, K., Reynolds, M., et~al.: Flamingo: a visual language
  model for few-shot learning. arXiv preprint arXiv:2204.14198  (2022)

\bibitem{bain2020condensed}
Bain, M., Nagrani, A., Brown, A., Zisserman, A.: Condensed movies: Story based
  retrieval with contextual embeddings. In: ACCV (2020)

\bibitem{Bain21}
Bain, M., Nagrani, A., Varol, G., Zisserman, A.: Frozen in time: A joint video
  and image encoder for end-to-end retrieval. In: Proc. ICCV (2021)

\bibitem{berg2022prompt}
Berg, H., Hall, S.M., Bhalgat, Y., Yang, W., Kirk, H.R., Shtedritski, A., Bain,
  M.: A prompt array keeps the bias away: Debiasing vision-language models with
  adversarial learning. arXiv preprint arXiv:2203.11933  (2022)

\bibitem{bogolin2022cross}
Bogolin, S.V., Croitoru, I., Jin, H., Liu, Y., Albanie, S.: Cross modal
  retrieval with querybank normalisation (June 2022)

\bibitem{brown2020language}
Brown, T., Mann, B., Ryder, N., Subbiah, M., Kaplan, J.D., Dhariwal, P.,
  Neelakantan, A., Shyam, P., Sastry, G., Askell, A., et~al.: Language models
  are few-shot learners. Advances in neural information processing systems
  \textbf{33},  1877--1901 (2020)

\bibitem{Carreira2017}
Carreira, J., Zisserman, A.: Quo vadis, action recognition? {A} new model and
  the {Kinetics} dataset. In: CVPR (2017)

\bibitem{Castro2022FitCLIPRL}
Castro, S., Heilbron, F.C.: Fitclip: Refining large-scale pretrained image-text
  models for zero-shot video understanding tasks. arXiv preprint
  arXiv:2203.13371  (2022)

\bibitem{castro2022fitclip}
Castro, S., Heilbron, F.C.: Fitclip: Refining large-scale pretrained image-text
  models for zero-shot video understanding tasks. arXiv preprint
  arXiv:2203.13371  (2022)

\bibitem{chen2021evaluating}
Chen, M., Tworek, J., Jun, H., Yuan, Q., Pinto, H.P.d.O., Kaplan, J., Edwards,
  H., Burda, Y., Joseph, N., Brockman, G., et~al.: Evaluating large language
  models trained on code. arXiv preprint arXiv:2107.03374  (2021)

\bibitem{chen2020uniter}
Chen, Y.C., Li, L., Yu, L., El~Kholy, A., Ahmed, F., Gan, Z., Cheng, Y., Liu,
  J.: Uniter: Universal image-text representation learning. In: European
  conference on computer vision. pp. 104--120. Springer (2020)

\bibitem{cheng2021improving}
Cheng, X., Lin, H., Wu, X., Yang, F., Shen, D.: Improving video-text retrieval
  by multi-stream corpus alignment and dual softmax loss (2021)

\bibitem{Croitoru21a}
Croitoru, I., Bogolin, S.V., Leordeanu, M., Jin, H., Zisserman, A., Albanie,
  S., Liu, Y.: Teachtext: Crossmodal generalized distillation for text-video
  retrieval. In: Proc. ICCV. IEEE (2021)

\bibitem{esmaeilpour2021zero}
Esmaeilpour, S., Liu, B., Robertson, E., Shu, L.: Zero-shot open set detection
  by extending clip. arXiv preprint arXiv:2109.02748  (2021)

\bibitem{fang2021clip2video}
Fang, H., Xiong, P., Xu, L., Chen, Y.: Clip2video: Mastering video-text
  retrieval via image clip. arXiv preprint arXiv:2106.11097  (2021)

\bibitem{fu2021violet}
Fu, T.J., Li, L., Gan, Z., Lin, K., Wang, W.Y., Wang, L., Liu, Z.: Violet:
  End-to-end video-language transformers with masked visual-token modeling.
  arXiv preprint arXiv:2111.12681  (2021)

\bibitem{Gabeur_2022_WACV}
Gabeur, V., Nagrani, A., Sun, C., Alahari, K., Schmid, C.: Masking modalities
  for cross-modal video retrieval. In: Proceedings of the IEEE/CVF Winter
  Conference on Applications of Computer Vision (WACV). pp. 1766--1775 (January
  2022)

\bibitem{gabeur2020multi}
Gabeur, V., Sun, C., Alahari, K., Schmid, C.: Multi-modal transformer for video
  retrieval. In: ECCV (2020)

\bibitem{Gaidon2013TemporalLO}
Gaidon, A., Harchaoui, Z., Schmid, C.: Temporal localization of actions with
  actoms. IEEE Transactions on Pattern Analysis and Machine Intelligence
  \textbf{35},  2782--2795 (2013)

\bibitem{ge2022bridgeformer}
Ge, Y., Ge, Y., Liu, X., Li, D., Shan, Y., Qie, X., Luo, P.: Bridgeformer:
  Bridging video-text retrieval with multiple choice questions. arXiv preprint
  arXiv:2201.04850  (2022)

\bibitem{goodwin2021semantically}
Goodwin, W., Vaze, S., Havoutis, I., Posner, I.: Semantically grounded object
  matching for robust robotic scene rearrangement. ICRA  (2022)

\bibitem{gu2021open}
Gu, X., Lin, T.Y., Kuo, W., Cui, Y.: Open-vocabulary detection via vision and
  language knowledge distillation. arXiv preprint arXiv:2104.13921  (2021)

\bibitem{Han22}
Han, T., Xie, W., Zisserman, A.: Temporal alignment networks for long-term
  video. In: CVPR (2022)

\bibitem{jia2021scaling}
Jia, C., Yang, Y., Xia, Y., Chen, Y.T., Parekh, Z., Pham, H., Le, Q.V., Sung,
  Y., Li, Z., Duerig, T.: Scaling up visual and vision-language representation
  learning with noisy text supervision. arXiv preprint arXiv:2102.05918  (2021)

\bibitem{DBLP:journals/corr/KingmaB14}
Kingma, D.P., Ba, J.: Adam: {A} method for stochastic optimization. In: Bengio,
  Y., LeCun, Y. (eds.) 3rd International Conference on Learning
  Representations, {ICLR} 2015, San Diego, CA, USA, May 7-9, 2015, Conference
  Track Proceedings (2015)

\bibitem{Korbar_2019_ICCV}
Korbar, B., Tran, D., Torresani, L.: Scsampler: Sampling salient clips from
  video for efficient action recognition. In: Proceedings of the IEEE/CVF
  International Conference on Computer Vision (ICCV) (October 2019)

\bibitem{krishna2017dense}
Krishna, R., Hata, K., Ren, F., Fei-Fei, L., Carlos~Niebles, J.:
  Dense-captioning events in videos. In: ICCV (2017)

\bibitem{li2022blip}
Li, J., Li, D., Xiong, C., Hoi, S.: Blip: Bootstrapping language-image
  pre-training for unified vision-language understanding and generation. arXiv
  preprint arXiv:2201.12086  (2022)

\bibitem{li2021align}
Li, J., Selvaraju, R., Gotmare, A., Joty, S., Xiong, C., Hoi, S.C.H.: Align
  before fuse: Vision and language representation learning with momentum
  distillation. Advances in Neural Information Processing Systems  \textbf{34}
  (2021)

\bibitem{ALBEF}
Li, J., Selvaraju, R.R., Gotmare, A.D., Joty, S., Xiong, C., Hoi, S.: Align
  before fuse: Vision and language representation learning with momentum
  distillation. In: NeurIPS (2021)

\bibitem{li2020oscar}
Li, X., Yin, X., Li, C., Zhang, P., Hu, X., Zhang, L., Wang, L., Hu, H., Dong,
  L., Wei, F., et~al.: Oscar: Object-semantics aligned pre-training for
  vision-language tasks. In: ECCV (2020)

\bibitem{eclipse}
Lin, Y.B., Lei, J., Bansal, M., Bertasius, G.: Eclipse: Efficient long-range
  video retrieval using sight and sound. arXiv preprint arXiv:2204.02874
  (2022)

\bibitem{Liu19a}
Liu, Y., Albanie, S., Nagrani, A., Zisserman, A.: Use what you have: Video
  retrieval using representations from collaborative experts. In: Proc. BMVC
  (2019)

\bibitem{Luo2021CLIP4Clip}
Luo, H., Ji, L., Zhong, M., Chen, Y., Lei, W., Duan, N., Li, T.: {CLIP4Clip}:
  An empirical study of clip for end to end video clip retrieval. arXiv
  preprint arXiv:2104.08860  (2021)

\bibitem{Miech2021ThinkingFA}
Miech, A., Alayrac, J.B., Laptev, I., Sivic, J., Zisserman, A.: Thinking fast
  and slow: Efficient text-to-visual retrieval with transformers. 2021 IEEE/CVF
  Conference on Computer Vision and Pattern Recognition (CVPR) pp. 9821--9831
  (2021)

\bibitem{miech17loupe}
Miech, A., Laptev, I., Sivic, J.: Learnable pooling with context gating for
  video classification. arXiv:1706.06905  (2017)

\bibitem{miech18learning}
Miech, A., Laptev, I., Sivic, J.: Learning a text-video embedding from
  incomplete and heterogeneous data. arXiv  (2018)

\bibitem{miech2019howto100m}
Miech, A., Zhukov, D., Alayrac, J.B., Tapaswi, M., Laptev, I., Sivic, J.:
  Howto100m: Learning a text-video embedding by watching hundred million
  narrated video clips. In: ICCV (2019)

\bibitem{mithun2019weakly}
Mithun, N.C., Paul, S., Roy-Chowdhury, A.K.: Weakly supervised video moment
  retrieval from text queries. In: Proceedings of the IEEE/CVF Conference on
  Computer Vision and Pattern Recognition. pp. 11592--11601 (2019)

\bibitem{mu2021slip}
Mu, N., Kirillov, A., Wagner, D., Xie, S.: Slip: Self-supervision meets
  language-image pre-training. arXiv preprint arXiv:2112.12750  (2021)

\bibitem{patrick2020support}
Patrick, M., Huang, P.Y., Asano, Y., Metze, F., Hauptmann, A., Henriques, J.,
  Vedaldi, A.: Support-set bottlenecks for video-text representation learning.
  arXiv preprint arXiv:2010.02824  (2020)

\bibitem{pirsiavash2014parsing}
Pirsiavash, H., Ramanan, D.: Parsing videos of actions with segmental grammars.
  In: Proceedings of the IEEE conference on computer vision and pattern
  recognition. pp. 612--619 (2014)

\bibitem{radford2021learning}
Radford, A., Kim, J.W., Hallacy, C., Ramesh, A., Goh, G., Agarwal, S., Sastry,
  G., Askell, A., Mishkin, P., Clark, J., Krueger, G., Sutskever, I.: Learning
  transferable visual models from natural language supervision. In: ICML (2021)

\bibitem{russakovsky2015imagenet}
Russakovsky, O., Deng, J., Su, H., Krause, J., Satheesh, S., Ma, S., Huang, Z.,
  Karpathy, A., Khosla, A., Bernstein, M., et~al.: Imagenet large scale visual
  recognition challenge. International journal of computer vision
  \textbf{115}(3),  211--252 (2015)

\bibitem{everything_at_once}
Shvetsova, N., Chen, B., Rouditchenko, A., Thomas, S., Kingsbury, B., Feris,
  R., Harwath, D., Glass, J., Kuehne, H.: Everything at once--multi-modal
  fusion transformer for video retrieval. arXiv preprint arXiv:2112.04446
  (2021)

\bibitem{sigurdsson2016hollywood}
Sigurdsson, G.A., Varol, G., Wang, X., Farhadi, A., Laptev, I., Gupta, A.:
  Hollywood in homes: Crowdsourcing data collection for activity understanding.
  In: ECCV (2016)

\bibitem{varol18_ltc}
Varol, G., Laptev, I., Schmid, C.: Long-term temporal convolutions for action
  recognition. IEEE Transactions on Pattern Analysis and Machine Intelligence
  \textbf{40}(6),  1510--1517 (2018)

\bibitem{adv_pertub}
Wang, J., Cherian, A.: Learning discriminative video representations using
  adversarial perturbations. In: Ferrari, V., Hebert, M., Sminchisescu, C.,
  Weiss, Y. (eds.) Computer Vision -- ECCV 2018. pp. 716--733. Springer
  International Publishing, Cham (2018)

\bibitem{wang2016temporal}
Wang, L., Xiong, Y., Wang, Z., Qiao, Y., Lin, D., Tang, X., Gool, L.V.:
  Temporal segment networks: Towards good practices for deep action
  recognition. In: European conference on computer vision. pp. 20--36. Springer
  (2016)

\bibitem{wang2016learning}
Wang, L., Li, Y., Lazebnik, S.: Learning deep structure-preserving image-text
  embeddings. In: CVPR (2016)

\bibitem{wang2021actionclip}
Wang, M., Xing, J., Liu, Y.: Actionclip: A new paradigm for video action
  recognition. arXiv preprint arXiv:2109.08472  (2021)

\bibitem{Wang_2021_CVPR}
Wang, X., Zhu, L., Yang, Y.: T2vlad: Global-local sequence alignment for
  text-video retrieval. In: Proceedings of the IEEE/CVF Conference on Computer
  Vision and Pattern Recognition (CVPR). pp. 5079--5088 (June 2021)

\bibitem{wu2019long}
Wu, C.Y., Feichtenhofer, C., Fan, H., He, K., Krahenbuhl, P., Girshick, R.:
  Long-term feature banks for detailed video understanding. In: Proceedings of
  the IEEE/CVF Conference on Computer Vision and Pattern Recognition. pp.
  284--293 (2019)

\bibitem{Wu_2021_CVPR}
Wu, C.Y., Krahenbuhl, P.: Towards long-form video understanding. In:
  Proceedings of the IEEE/CVF Conference on Computer Vision and Pattern
  Recognition (CVPR). pp. 1884--1894 (June 2021)

\bibitem{wu2022memvit}
Wu, C.Y., Li, Y., Mangalam, K., Fan, H., Xiong, B., Malik, J., Feichtenhofer,
  C.: Memvit: Memory-augmented multiscale vision transformer for efficient
  long-term video recognition. arXiv preprint arXiv:2201.08383  (2022)

\bibitem{xu2021videoclip}
Xu, H., Ghosh, G., Huang, P.Y., Okhonko, D., Aghajanyan, A., Metze, F.,
  Zettlemoyer, L., Feichtenhofer, C.: Videoclip: Contrastive pre-training for
  zero-shot video-text understanding. arXiv preprint arXiv:2109.14084  (2021)

\bibitem{xu2016msr}
Xu, J., Mei, T., Yao, T., Rui, Y.: Msr-vtt: A large video description dataset
  for bridging video and language. In: CVPR (2016)

\bibitem{Yan2021VideoTextPW}
Yan, R., Shou, M.Z., Ge, Y., Wang, A., Lin, X., Cai, G., Tang, J.: Video-text
  pre-training with learned regions. ArXiv  \textbf{abs/2112.01194} (2021)

\bibitem{Yang_2021_ICCV}
Yang, A., Miech, A., Sivic, J., Laptev, I., Schmid, C.: Just ask: Learning to
  answer questions from millions of narrated videos. In: Proceedings of the
  IEEE/CVF International Conference on Computer Vision (ICCV). pp. 1686--1697
  (October 2021)

\bibitem{yu2018joint}
Yu, Y., Kim, J., Kim, G.: A joint sequence fusion model for video question
  answering and retrieval. In: ECCV (2018)

\bibitem{yue2015beyond}
Yue-Hei~Ng, J., Hausknecht, M., Vijayanarasimhan, S., Vinyals, O., Monga, R.,
  Toderici, G.: Beyond short snippets: Deep networks for video classification.
  In: Proceedings of the IEEE conference on computer vision and pattern
  recognition. pp. 4694--4702 (2015)

\bibitem{zhai2021scaling}
Zhai, X., Kolesnikov, A., Houlsby, N., Beyer, L.: Scaling vision transformers.
  arxiv. arXiv preprint arXiv:2106.04560  (2021)

\bibitem{zhao2019long}
Zhao, Z., Zhang, Z., Xiao, S., Xiao, Z., Yan, X., Yu, J., Cai, D., Wu, F.:
  Long-form video question answering via dynamic hierarchical reinforced
  networks. IEEE Transactions on Image Processing  \textbf{28}(12),  5939--5952
  (2019)

\bibitem{Zhou2021LearningTP}
Zhou, K., Yang, J., Loy, C.C., Liu, Z.: Learning to prompt for vision-language
  models. arXiv preprint arXiv:2109.01134  (2021)

\end{thebibliography}
